\pdfoutput=1
\PassOptionsToPackage{table}{xcolor}

\documentclass[11pt]{article}

\usepackage[final]{acl}

\usepackage[table]{xcolor}

\usepackage{times}
\usepackage{listings}
\usepackage{latexsym}
\usepackage{todonotes}
\usepackage{tabularx}
\usepackage{booktabs}
\usepackage{enumitem}
\usepackage{geometry}
\usepackage{makecell}
\usepackage{algorithm}
\usepackage{algpseudocode}
\usepackage{graphicx}
\usepackage{adjustbox}
\usepackage{amsbsy}
\usepackage[acronym]{glossaries}
\usepackage{multirow}

\usepackage{todonotes}

\newcommand{\NumFilteredHumanAnnotations}{453 }

\newcommand{\iiutod}{\textsc{IndirectRequests}}

\newcommand{\NumValidAndTestSamples}{330 }

\newcommand{\ProxyEvaluator}[1][singular]{%
    \ifthenelse{\equal{#1}{plural}}{proxy evaluators}{proxy evaluator}%
}
\glsdisablehyper
\newacronym{iiu}{IUR}{Indirect User Request} 
\newacronym{llm}{LLM}{large language model}
\newacronym{sgd}{\textsc{SGD}}{Schema Guided Dialog}
\newacronym{cot}{CoT}{Chain-of-Thought}
\newacronym{dst}{DST}{Dialogue State Tracking}
\newacronym{nli}{NLI}{Natural Language Inference}
\newacronym{nlu}{NLU}{Natural Language Understanding}
\newacronym{icl}{ICL}{in-context learning}
\newacronym{iiu-tod}{\textsc{IndirectRequests}}{\textsc{IndirectRequests}}

\usepackage[T1]{fontenc}

\usepackage[utf8]{inputenc}

\usepackage{microtype}

\newcommand{\appropriateness}{\textsc{Appropriateness}}
\newcommand{\unamb}{\textsc{Unambiguity}}
\newcommand{\knowledge}{\textsc{World-Understanding}}

\title{Making Task-Oriented Dialogue Datasets More Natural by Synthetically Generating Indirect User Requests}

\author{
  Amogh Mannekote \\
  University of Florida\thanks{Work done during an internship at Amazon} \\
  \And
  Jinseok Nam \\
  Amazon \\
  \And
  Ziming Li \\
  Amazon \\
  \AND
  Jian Gao \\
  Amazon \\
  \And
  Kristy Elizabeth Boyer \\
  University of Florida \\
  \And
  Bonnie J. Dorr \\
  University of Florida
}

\begin{document}
\maketitle
\begin{abstract}

\glspl{iiu}, such as "It's cold in here" instead of "Could you please increase the temperature?" are common in human-human task-oriented dialogue and require world knowledge and pragmatic reasoning from the listener.
While \glspl{llm} can handle these requests effectively, smaller models deployed on virtual assistants often struggle due to resource constraints.
Moreover, existing task-oriented dialogue benchmarks lack sufficient examples of complex discourse phenomena such as indirectness.
To address this, we propose a set of linguistic criteria along with an \gls{llm}-based pipeline for generating realistic \glspl{iiu} to test \gls{nlu} and \gls{dst} models before deployment in a new domain.
We also release \iiutod, a dataset of \glspl{iiu} based on the \gls{sgd} corpus, as a comparative testbed for evaluating the performance of smaller models in handling indirect requests.

\end{abstract}

\glsresetall

\section{Introduction}

\begin{figure}
    \centering
    \includegraphics[width=\columnwidth]{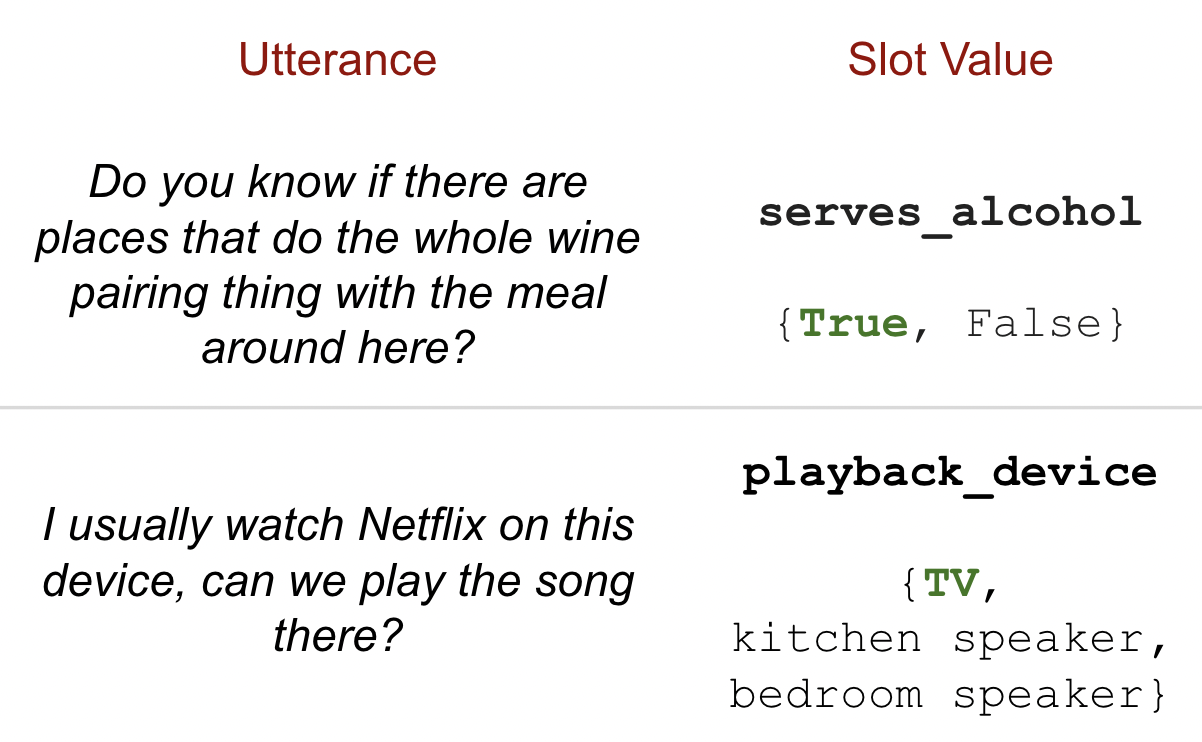}
    \caption{Two settings are illustrated for 
    \glspl{iiu}: restaurant-reservation and home-automation.}
    \label{fig:iiu-example}
    \vspace*{-.2in}
\end{figure}

Non-literal, indirect utterances are common in human-human task-oriented dialogue and require pragmatic understanding and world knowledge for successful interpretation (e.g., \textit{``It's cold in here''} instead of \textit{``Could you please increase the temperature?''}) \citep{briggs2017strategies}. This phenomenon is a key area of interest in discourse pragmatics \citep{blum2011discourse, schegloff1999discourse}, supported by theoretical frameworks such as Grice's maxims \citep{grice1975logic} and RST \citep{mann1988rhetorical}. Figure \ref{fig:iiu-example} illustrates two instances of \glspl{iiu}.

Despite the prevalence of indirect utterances in everyday discourse and the human-level \gls{nlu} performance demonstrated by state-of-the-art \glspl{llm} like GPT-4 \citep{achiam2023gpt}, current virtual assistants struggle to handle such utterances seamlessly \citep{mavrina_alexa_2022}. This can be attributed, in part, to the high computational cost associated with using state-of-the-art, large models for inference \citep{samsi2023words, sardana2023beyond}. A common workaround is to employ smaller, cost-effective, task-specific models \citep{hsieh2023distilling}. However, this approach often compromises the generalizability and robustness offered by larger models.



Over the years, several benchmark datasets for task-oriented dialogue, such as MultiWOZ \citep{budzianowski2018multiwoz}, \gls{sgd} \citep{rastogi2020towards}, and FRAMES \citep{asri2017frames}, have been curated by the dialogue systems community. However, these datasets have two key limitations that hinder their effectiveness in training smaller \gls{nlu} models. First, their static nature and limited domain coverage make it difficult to evaluate \gls{nlu} or \gls{dst} models in new domains. Second, the controlled laboratory settings in which these datasets are crowdsourced lead to a distributional mismatch between the benchmark datasets and ``in-the-wild'' utterances \cite{zarcone2021small}.

\section{Schema-Guided Dialogue} \label{sec:notation}

To bridge this distributional gap, we 
present
an \gls{llm}-based data generation pipeline to scalably generate \glspl{iiu} for a new task-oriented dialogue domain. 
Our
work 
makes the following contributions:
\begin{enumerate}
\item 
We develop a set of
linguistic criteria to formalize the concept of what constitutes an indirect user request in a task-oriented dialogue setting. 
\item We develop a 
 pipeline to collect gold-labelled \glspl{iiu},
using an \gls{llm} to generate a noisy, seed \gls{iiu} dataset, followed by
crowd-sourced filtering and correction to increase quality. 
\item We publicly release \iiutod, 
a dataset of \glspl{iiu} collected 
through the process above,
using the schemas from the \gls{sgd} dataset. 
We aim for it to serve as a testbed for both researchers and practitioners interested in evaluating model robustness.
\item To circumvent the need for collecting expensive human labels for a new domain, we report results over various ``proxy'' models for \textit{automatically} evaluating the quality of \glspl{iiu} according to our linguistic criteria.
\item Finally, 
we empirically demonstrate the increased difficulty of the \glspl{iiu} by showing 
that the performance of 
a T5-based \citep{roberts2019exploring} \gls{dst} model significantly degrades when applied on \iiutod\ utterances
as compared to their counterparts from \gls{sgd}.  
\end{enumerate}

Before outlining the linguistic criteria, we first describe the paradigm of ``schema-guided dialogue'' since it serves as the basis for the task formulation.

\begin{figure}[t]
    \centering
    \includegraphics[scale=0.9]{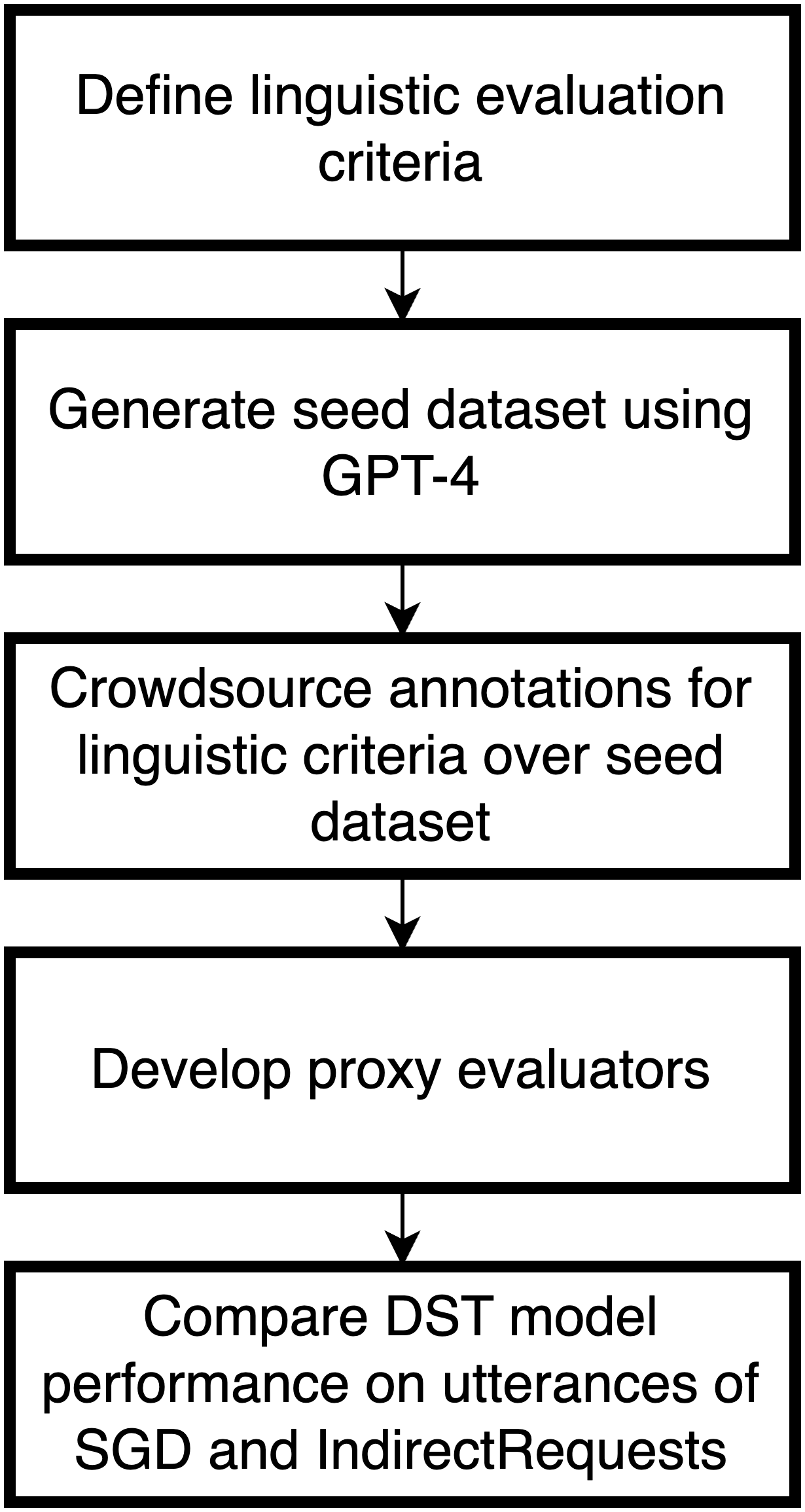}
    \caption{The five-stage \gls{iiu} generation pipeline.}
    \label{fig:pipeline}
    \vspace*{-.2in}
\end{figure}

A long-standing goal in task-oriented dialogue research has been zero-shot transfer of critical modules such as the \gls{nlu} and \gls{dst} to previously unseen domains and backend APIs \citep{mehri2022lad}.
    To achieve this goal, we need a way to represent new domains and APIs in a format that can be fed to a machine learning model. In addition, it helps if the representation is made as succinct to achieve both conceptual simplicity and human readability \citep{Mannekote2023ExploringUI}.
    A ``dialogue schema'' is 
    any structured format that performs this role of describing a domain that a dialogue system will operate in.

\begin{figure}[hptb]
    \centering
    \includegraphics[width=0.7\columnwidth]{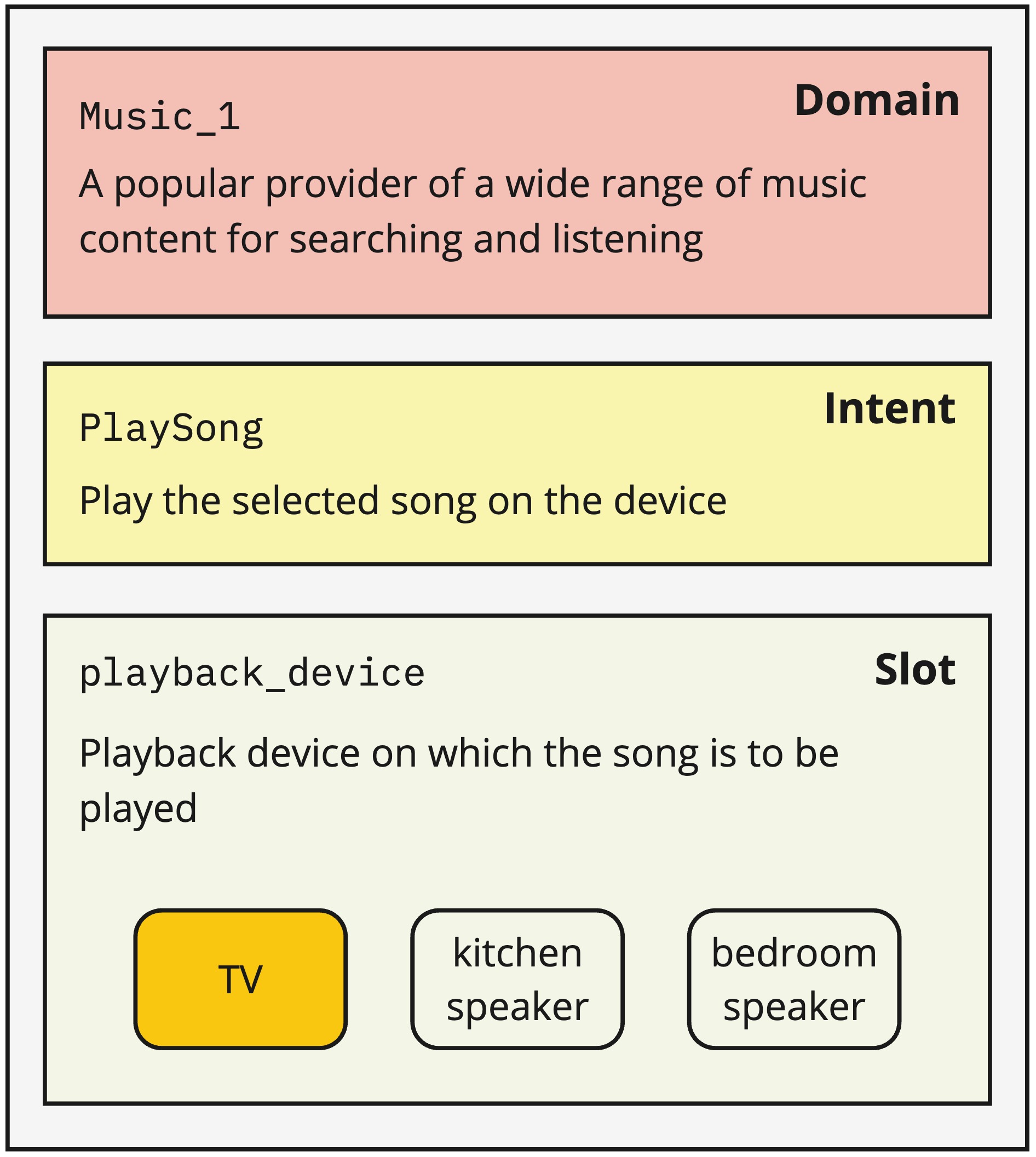}
    \caption{We illustrate a dialogue schema in the music service domain, with an intent to play music and a slot for selecting a playback device (e.g., TV, kitchen speaker, bedroom speaker). Our approach generates an indirect utterance based on a specified slot value, such as 'TV.'}
    \label{fig:dial-schema}
    \vspace*{-.2in}
\end{figure}

To facilitate shared tasks, \citet{rastogi2020towards} 
    formally introduce 
    the paradigm of ``schema-guided dialogue'' 
    alongside a benchmark corpus: the \gls{sgd} dataset.
Their schemas
(shown in Figure \ref{fig:dial-schema}) factor each task-oriented dialogue domain into its constituent \textit{intents} and \textit{slots}.

Consider a \texttt{Movie} domain 
consisting of
two intents: \texttt{RentMovie} and \texttt{BuyTickets}. 
To satisfy each intent, the user needs to fill a set of slots. Slots can be 
considered analogous to
query fields for an API call. For example, to fulfill the \texttt{BuyTickets} intent, the schema can demand that the \texttt{NumPeople}, \texttt{MovieName}, and \texttt{Date} slots be filled.
A crucial aspect 
of \gls{sgd}'s schemas
is their
use of one-line natural language descriptions to describe the domain, intents, and slots. This design
allows language models to make effective use of the schemas.



\section{Linguistic Criteria} \label{sec:evaluation-rubric}

We propose evaluating indirectness using three linguistic criteria: \appropriateness{}, \unamb{}, and \knowledge{}. For each criterion, Table \ref{tab:evaluation-metrics} shows examples of utterances that fall on the extreme ends of the rating scales. Note that each of the three labels carries a more precise meaning as compared to their freer usage in everyday language. 

\begin{table*}[ht]
    \centering
    \small
    \begin{tabularx}{\textwidth}{>{\raggedright\arraybackslash}p{2.8cm} >{\raggedright\arraybackslash}X >{\raggedright\arraybackslash}X >{\raggedright\arraybackslash}X}
        \toprule
        \textbf{Linguistic Criterion} & \textbf{High-Scoring Utterance} & \textbf{Low-Scoring Utterance} & \textbf{Justification} \\\\
        \midrule
        \textbf{\appropriateness{}} & \textit{I'm looking for tickets that I can exchange or refund in case of a change in plan.} & \textit{I'd like to order a sandwich.} & The low-scoring example is nonsensical in the context of buying a bus ticket. \\\\
        \textbf{\unamb{}} & \textit{I'm looking for tickets that I can exchange or refund in case of a change in plan.} & \textit{I’m looking for tickets that give me additional benefits.} & The term ``additional benefits'' is ambiguous as it can refer to either \textit{Flexible} or \textit{Economy Extra}. \\\\
        \textbf{\knowledge{}} & \textit{Do you know of any Michelin star restaurants in the area that offer a unique dining experience?} & \textit{I'm looking to treat myself to a luxurious meal with the highest quality ingredients, so I'd like to find a restaurant like that} & ``Michelin star'' demonstrates more in-depth world knowledge as opposed to ``luxurious meal.'' \\\\
        \bottomrule
    \end{tabularx}
    \caption{Criteria to Evaluate \glspl{iiu} are provided with two accompanying example utterances: one that is high-scoring on that criterion, and another that is low-scoring. \label{tab:evaluation-metrics}}
    \vspace*{-.2in}
\end{table*}

\paragraph{\appropriateness{}.}
The \appropriateness{} criterion seeks to ensure that an \gls{iiu} does not sound out of place in the real-world context it is being uttered in.
For instance, the utterance \textit{``I'd like to order a sandwich''} would be completely irrelevant in a setting where the user is trying to book bus tickets.
In contrast, the utterance \textit{``I want to go somewhere''} would be relevant.

\paragraph{\unamb{}.}
The \unamb{} criterion is designed 
to ensure that a generated \gls{iiu} entails the target slot value, not any 
of the remaining candidate slot values.
For instance, 
a flight-booking scenario 
includes a ``seating class'' slot 
with 
values such as ``Economy,'' ``Premium Economy,'' ``Business,'' and ``First Class.'' 
Thus, 
the utterance \textit{``I'm looking to book a luxurious seat on the flight''} is ambiguous, since the user could arguably be referring to
any of these values.

\begin{figure*}[hptb]
    \centering
    
    \includegraphics[scale=0.4]{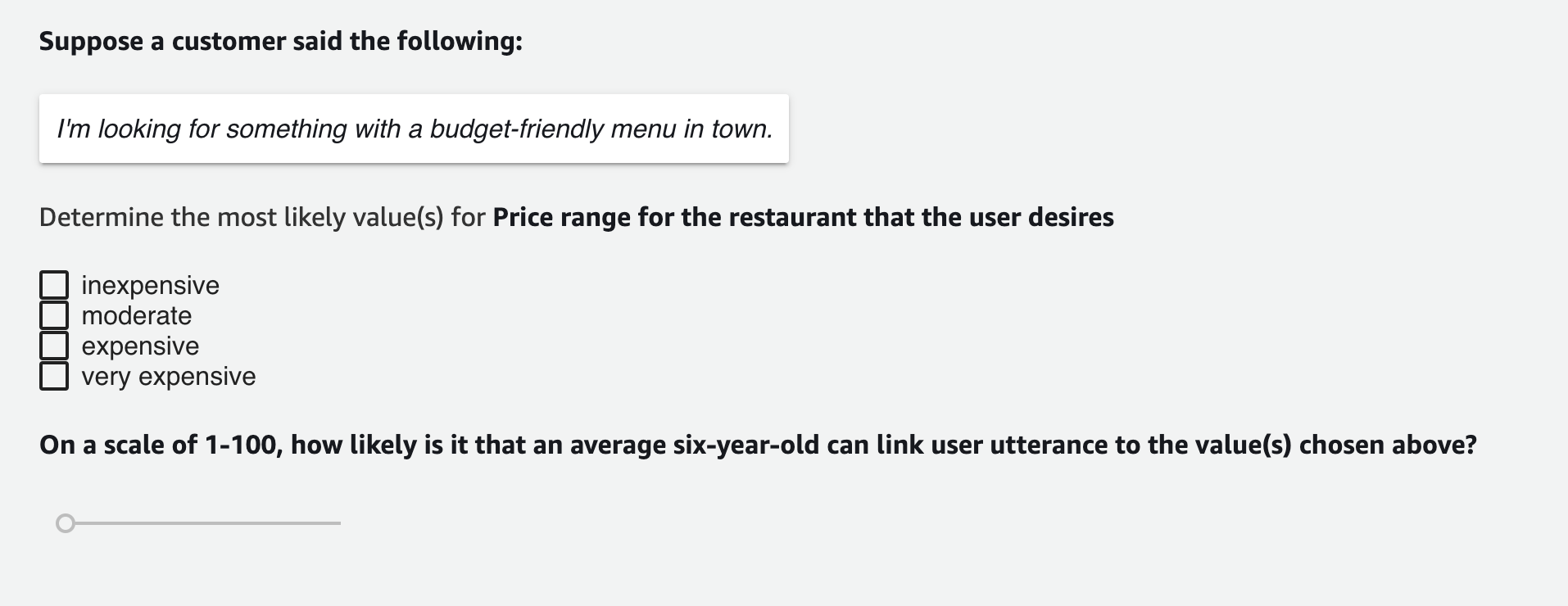}
    \caption{
        The M-Turk crowdsourcing interface for collecting human annotations over the seed dataset contains two form elements.
        The first assesses the \unamb{} in the generated utterance, ensuring that it entails only the target slot value.
        The second assesses the \knowledge{} criterion, leveraging a slider to rate the likelihood that an average six-year-old could correctly infer the target slot value. The latter is an intuitive proxy to measure the complexity of world understanding required to interpret the utterance.
    }
    \label{fig:crowdsourcing-interface}
    \vspace*{-.2in}
\end{figure*}

\paragraph{\knowledge{}.}
The \knowledge{} criterion is intended to be a measure of the degree of world understanding required by the listener to draw the connection between an \gls{iiu} and the user's intended target slot value. For example, when filling the \textit{destination-country} slot in a trip-booking scenario, the utterance \textit{``I'm looking to book a ticket to an African country''} can refer to values such as ``Nigeria'' or ``Egypt'' but not ``India.''

\section{The \iiutod\ Dataset} \label{sec:bootstrapping-seed-corpus}

The goal of \gls{iiu} generation is to take a domain, a domain schema (containing a user intent and a list of possible slot values), and a target slot value as inputs and output an \gls{iiu}. The \gls{iiu}, on its part, is expected to adhere to certain ``linguistic criteria'' to be valid.

Given a set of linguistic criteria for evaluating the quality of text samples, there are two broad approaches to crowdsource a dataset: (1) present real-world scenarios to crowdworkers and ask them to compose corresponding \glspl{iiu} in an open-ended manner, or (2) provide pre-generated \glspl{iiu} and ask crowdworkers to rate the quality of each \gls{iiu} on a numerical scale reflecting the desired linguistic criteria. While the first approach demands crowdworkers to apply the provided linguistic framework, exhibit creativity, and possess proficient writing skills, rendering it expensive, the second approach involves the simpler task of evaluating existing utterances. Therefore, we generate a large number of (potentially noisy) \glspl{iiu} using a combination of GPT-3.5 \cite{brown2020language} and GPT-4 models from OpenAI, and then ask crowdworkers to rate their quality based on our linguistic criteria.

\subsection{Generating the Seed Dataset}
In order to prompt an \gls{llm} for a task, we need a prompting strategy (operationalized using what is commonly referred to as a ``prompt template''). While prompt engineering is an open-ended process, we follow guiding principles such as making instructions specific and detailed, including high-quality in-context examples, and exploiting strategies like \gls{cot} \citep{wei2022chain} to improve output quality.
We use \gls{cot} prompting \citep{wei2022chain} to generate \glspl{iiu}, as it has been shown to improve performance on NLP tasks involving reasoning, such as ours. This technique breaks down a problem into intermediate steps. For our task, we first generated a set of ``interesting facts'' about the target slot value in the given situation context, and then generated the final \glspl{iiu} conditioned on those facts.
Therefore, this strategy was employed to scale up and generate a comprehensive seed dataset consisting of \NumFilteredHumanAnnotations \glspl{iiu}.

\subsection{Crowdsourcing Human Labels} \label{sec:crowdsourcing}

Manual inspection of the \glspl{iiu} in the seed dataset reveals
considerable variation in quality, suggesting 
a need for refinement
before utilizing them as gold-labeled data for evaluation.
To address this, we set up a crowdsourcing pipeline using Amazon Mechanical Turk (M-Turk) to have crowdworkers rate the quality of the candidate \glspl{iiu} in accordance with our linguistic criteria.

There are two key considerations for
developing the crowdsourcing interface: 
1) to optimize annotator efficiency (reducing the time and effort required per evaluated sample) and 2) to maximize inter-annotator agreement.
We observe
that the variation in the unannotated seed dataset is predominantly along the criteria of \unamb{} and \knowledge{}. Only a negligible number of instances were deemed irrelevant based on the \appropriateness{} criteria. Consequently, we streamline the interface to include two primary components, one each for evaluating \unamb{} and \knowledge{}.

\paragraph{\unamb{} Annotation.} To collect labels for the \unamb{} criterion, we instruct the annotators to select all the slot values (zero or more) that they think are entailed by the utterance using a multiple choice checkbox (the annotator can check one or more boxes). We 
design this form element as
a binary yes/no question to avoid posing the question in a leading way.  Multiple selections by an annotator imply the utterance fails to meet the \unamb{} criterion.

\paragraph{\knowledge{} Annotation.} For the \knowledge{} criterion, we ask annotators to engage in a thought experiment where they adopt the perspective of a six-year-old child. This approach aims to assess whether a connection between the utterance and selected slot values would be discernible to a child of that age.
We arrived at this unique framing after several iterations of refining the question. Initially, we asked annotators directly to rate the ``complexity'' involved in making the connection. However, we recognized that the concept of ``complexity'' is highly subjective and can vary significantly among individuals.
To standardize the perception of complexity and reduce variability among annotators, we anchor our assessment to a child's level of understanding. This approach aims to provide a consistent benchmark, despite the diverse cognitive abilities typically present at that age range.

\subsection{Dataset Splits}
Based on the crowdsourced labels for both \unamb{} and \knowledge{}, we curate the \iiutod\ dataset and release it for public use.\footnote{URL hidden for peer review.}
In going from the ``raw'' crowdsourced samples to the dataset, we split the dataset and systematically create labels for each sample
for both \unamb{} and \knowledge{} criteria.
While splitting \iiutod\ into train, validation, and test sets, we split our samples based on same lines on which the services 
are split across the \gls{sgd} dataset. 
This alignment with the \gls{sgd} dataset splits is intended to aid future work that might need to compare our results with previous work reporting on \gls{sgd}.

\begin{table}[h!]
\centering
\begin{tabular}{ccc}
\toprule
\textbf{Train} & \textbf{Validation} & \textbf{Test} \\ \midrule
123 & 136 & 194 \\ \bottomrule
\end{tabular}
\caption{Number of samples in each split of {\iiutod}}
\label{table:samples}
\end{table}

\begin{table*}[] 
\centering
\begin{tabular}{@{}ccrrrrr@{}}
\toprule
\multirow{3}{*}{\textbf{Criterion}} &
  \multicolumn{6}{c}{\textbf{Model}} \\ \cline{2-7} 
 &
  \multirow{2}{*}{\textbf{\begin{tabular}[c]{@{}c@{}}Small\\ LM (\textless{}1B)\end{tabular}}} &
  \multicolumn{2}{c}{\textbf{GPT (3-shot)}} &
  \multicolumn{3}{c}{\textbf{Llama 2 (3-shot)}} \\
 &
   &
  \multicolumn{1}{l}{\textbf{GPT-3.5}} &
  \multicolumn{1}{l}{\textbf{GPT-4}} &
  \multicolumn{1}{l}{\textbf{7B}} &
  \multicolumn{1}{l}{\textbf{13B}} &
  \multicolumn{1}{l}{\textbf{70B}} \\ \hline
\multicolumn{1}{l}{\begin{tabular}[c]{@{}l@{}}\unamb{}\\ (Accuracy)\end{tabular}} &
  \multicolumn{1}{r}{\begin{tabular}[c]{@{}r@{}}0.35\(^{*}\)\\ (nli-deberta)\end{tabular}} &
  0.73 &
  0.84\(^{\dagger}\) &
  0.5 &
  0.69\(^{\ddagger}\) &
  0.22 \\
\multicolumn{1}{l}{\begin{tabular}[c]{@{}l@{}}\knowledge{}\\ (Pearson correlation)\end{tabular}} &
  \multicolumn{1}{r}{\begin{tabular}[c]{@{}r@{}}0.22\(^{*}\)\\ (ms-marco)\end{tabular}} &
  0.15 &
  0.34\(^{\dagger}\) &
  0.16 &
  0.19\(^{\ddagger}\) &
  0.18 \\ \hline
\end{tabular}
\caption{Evaluation results are computed from a single run with proxy evaluators against crowdworker annotations on the combined validation and test splits of \iiutod, which contain a total of \NumValidAndTestSamples samples. Performance symbols indicate the best-performing models within specific categories. \(^\star\) denotes the best performance in the zero-shot (small LM) category, \(^\dagger\) marks the best performance in the proprietary OpenAI \gls{llm} category, and \(^\ddagger\) signifies the top performer among the Llama 2 models \cite{touvron2023llama}.
\label{tab:proxy-evaluator-results}}
\vspace*{-.2in}
\end{table*}

\section{Proxy Evaluation of Linguistic Criteria} \label{sec:proxy-evaluators}


We 
perform an automated, 
proxy evaluation of the \glspl{iiu} generations 
due to the impracticality of manually evaluating the large number of samples and models.
In this section, we define the proxy evaluation task formulations and present baseline results using zero-shot and few-shot prompting strategies.
We define two proxy evaluation tasks, corresponding to the \unamb{} and \knowledge{} criteria, respectively.

    
\paragraph{\unamb{}.}
We frame proxy evaluation of \unamb{} as a multi-class classification problem with $N_i + 1$ classes, where $N_i$ is the number of possible slot values for the given slot $i$. 
We 
add an extra class corresponding to the case where the ground truth (from the crowdsourcing step) is ambiguous. For model comparison, we report the accuracy over all samples in the test split.


\paragraph{\knowledge{}.}
We define the proxy evaluation of \knowledge{} as predicting the level of world knowledge required to infer the intended slot value from an utterance as a continuous value ranging from 1 to 10. This approach aligns with the methodology used in our crowdsourcing stage, where judgments about knowledge depth were made using a 1-100 scale slider. Performance is quantified by calculating the sum of squared errors between predicted and actual values (after normalizing both sets of values).



\subsection{Proxy Evaluation Results}
We split the proxy evaluation models into three categories: small language models (fewer than 1B 
parameters), proprietary large language models from OpenAI (gpt-3.5-turbo and gpt-4-0125-preview), and open-source Llama 2 language models (7B, 13B, and 70B). Table \ref{tab:proxy-evaluator-results} shows the performance of the proxy evaluators on the test split against the ground truth obtained through crowdsourcing.

\paragraph{Small LMs.}
For the small LM category, we employ BERT-based models in a zero-shot setup. For the \unamb{} criterion, we frame the evaluation as $k$ \gls{nli} problems, where $k$ is the number of possible slot values. Each problem considers the candidate \gls{iiu} as the premise and a possible slot value as the hypothesis. We use a BERT-based \gls{nli} model\footnote{nli-deberta-v3-small} to obtain entailment scores and return the argmax score. If the maximum score is below 0.3, we deem the \gls{iiu} ambiguous for that slot. For \knowledge{}, we use ms-marco-MiniLM-L-6-v2\footnote{https://huggingface.co/microsoft/ms-marco-MiniLM-L-6-v2}, fine-tuned on MS MARCO for passage ranking. We concatenate the \gls{iiu} with the knowledge context, score the sequence using the model, and assign a \knowledge{} rating of 10 if the the score exceeds 0.5 and 0 otherwise.

\paragraph{Proprietary LLMs.}
For the proprietary LLMs from OpenAI, we use the models in a few-shot setup, providing a few examples of \gls{iiu}s labeled as either ambiguous or unambiguous (for \unamb{}), or knowledgeable or not knowledgeable (for \knowledge{}). We then query the model with the test \gls{iiu} and knowledge context (if applicable) and take the model's output as the prediction.

\paragraph{Open-Source LLMs.}
For the open-source Llama 2 models (7B, 13B, and 70B), we use a similar few-shot setup as we did with the proprietary LLMs.
Table \ref{tab:proxy-evaluator-results}, summarizes these results. 

While achieving high inter-annotator agreement (IAA) for subjective measures like \knowledge{} and \unamb{} is inherently challenging, as evidenced by prior work showing human annotators struggling to exceed 30\% IAA for related subjective criteria in NLG tasks \cite{karpinska-etal-2021-perils}, we find that \gls{llm}-based proxy evaluation models, particularly GPT-3.5 and GPT-4, demonstrate considerable agreement with human raters for our task. Nonetheless, there remains scope for further boosting performance through additional prompt engineering and experimentation with adaptive strategies for selecting in-context examples. The prompts used for training both proprietary and open-source \gls{llm} proxy evaluator models are provided in Appendix \ref{app:prompts-for-proxy-evaluator}.


\section{Automated \gls{iiu} Generation} \label{sec:knowledge-conditioned-iiu-gen}


\begin{table*}[t]
\begin{adjustbox}{scale=0.8}
\begin{tabular}{@{}m{4cm}m{4.5cm}m{11cm}@{}}
\toprule
\begin{tabular}[c]{@{}l@{}}\textbf{Indirection}\\\textbf{Strategy}\end{tabular} &
  \textbf{Intent-Slot-Value} &
  \textbf{Sample \gls{iiu}} \\ \midrule
Simple Elaboration &
  \begin{tabular}[c]{@{}l@{}}RentMovie\\(subtitles = None)\end{tabular} &
  ``I prefer watching films in their native language \textbf{without any language barriers}.'' \\
Justification &
  \begin{tabular}[c]{@{}l@{}}GetRide\\(shared\_ride = True)\end{tabular} &
  ``I usually like sharing the ride with someone else \textbf{to reduce carbon footprint}...'' \\
Hyponym Swap &
  \begin{tabular}[c]{@{}l@{}}SearchEvents\\(type = Music)\end{tabular} &
  ``Is there a festival happening around with \textbf{pop}, \textbf{country} or \textbf{hip-hop} artists performing?'' \\
Synonym Swap &
  \begin{tabular}[c]{@{}l@{}}RentMovie\\(subtitles = Mandarin)\end{tabular} &
  ``I've got a bunch of friends coming over who are more comfortable with \textbf{Simplified Chinese}. Can you find me movies...'' \\
Small Talk &
  \begin{tabular}[c]{@{}l@{}}FindApartment\\(pets\_allowed = True)\end{tabular} &
  ``I'm looking for a place where my dog is allowed to come along. \textbf{He's so cute and he doesn't shed as much as you think!}'' \\
  \bottomrule
\end{tabular}
\end{adjustbox}
\caption{From the generated \glspl{iiu}, we identify five main indirection strategies (Simple Elaboration, Justification, Hyponym Swap, Synonym Swap, and Small Talk).\label{tab:qualitative-results}}
\vspace*{-.2in}
\end{table*}


Under ideal conditions, we would use as small an \gls{llm} as possible to generate high-quality \glspl{iiu}.
We report the quality of the generated \glspl{iiu} generated using smaller, open-source \glspl{llm} (Llama 2) in Table \ref{fig:llama-generation-results}. The prompt used to generate the \glspl{iiu} is given in Appendix \ref{app:iiu-generation-prompt}.


\subsection{Indirection Strategies} \label{sec:qualitative-analysis}
Along with reporting quantitative metrics from our proxy evaluators, we also perform a bottom-up content analysis 
to develop a richer understanding of the specific ``indirection strategies'' that the \glspl{llm} employ to transform the slot schema into \glspl{iiu}.
During analysis, one of the authors excluded 
those samples for which the \gls{iiu} either very evidently does not entail the target slot value or the slot value is mentioned
verbatim,
violating the \unamb{} criterion.

We identify five main indirection strategies from our content analysis (see Table \ref{tab:qualitative-results}).
    \textbf{Simple Elaboration} performs a simple replacement of the slot value with a longer phrase meaning the same thing. Simple Elaborations do not leverage non-trivial world knowledge.
    \textbf{Justification} 
    offers a real-world reason for choosing a particular slot value.
    A \textbf{Hyponym Swap} involves replacing the slot value with its hyponym (the replacement is a more specific instance or subtype of the original term).
    Similarly, a \textbf{Synonym Swap} replaces the slot value with a synonym.
    The final strategy, \textbf{Small Talk}, involves padding the utterance with information that is not strictly informational to the task. While this is not strictly an indirection strategy, it can serve to complement another indirection strategy by making it sounds more realistic.

\section{Extrinsic Evaluation} \label{sec:extrinsic-evaluation}
While intrinsic, automated evaluations provide valuable insights, we further assess the practical implications of \iiutod\ through extrinsic evaluation, measuring the performance degradation of a widely-adopted \gls{dst} model on our dataset compared to its performance on the canonical \gls{sgd} corpus. This approach aligns with established practices in the dialogue systems literature, where \gls{nlu} model performance is extensively evaluated in isolation, as it critically impacts downstream dialogue policy learning and response generation in modular architectures.

Our objective is not to conduct an end-to-end evaluation of dialogue systems, but to specifically evaluate \gls{nlu} performance. By providing a relative comparison against the commonly referenced \gls{sgd} corpus, we aim to highlight the increased parsing difficulty posed by \iiutod\ utterances, rather than claiming they present challenges to state-of-the-art models, including \gls{llm}-based ones. This targeted evaluation allows us to isolate and characterize the unique aspects of our dataset, contributing to a more comprehensive understanding of \gls{nlu} model capabilities and limitations.


Since the \gls{dst} model we use is trained on context window lengths of 3, the dialogue contexts in all samples are also set to 3.
Table \ref{tab:sgd-vs-iu} shows a comparison between the model performance over the original samples and the samples using the generated \glspl{iiu} based on a total of \NumValidAndTestSamples samples. 



To 
fairly compare the results of any \gls{nlu} model over \gls{sgd} and \iiutod\
during extrinsic evaluation, we only use a subset of \gls{sgd} that satisfies the following conditions:
\begin{enumerate}[itemsep=2pt, parsep=0pt] 
    \item 
    user request must be about a categorical slot
    \item 
    speaker of the latest utterance in the dialogue context 
    must be the user and not the system
    \item 
    dialogue act of the latest utterance should be ``inform'' (as opposed to ``request'' utterances, which is out of scope for our work)
    \item 
    user utterance includes only a single slot-value pair (since our \gls{iiu} generation method does not accommodate more than one slot-value pair per \gls{iiu})
\end{enumerate}


\begin{table}[ht]
    \centering
    \begin{tabular}{lcc}
        \toprule
        \textbf{Base Model} & \textbf{\gls{sgd}} & \textbf{\iiutod} \\
        \midrule
        T5 & 0.512 & 0.133 \\
        \bottomrule
    \end{tabular} 
    \caption{Slot accuracies are computed for a T5-based state-of-the-art dialogue state tracking model on samples from both the original \gls{sgd} dataset and the \iiutod. The DST model performance on \iiutod\ shows a significant degradation. \label{tab:sgd-vs-iu}}
    \vspace*{-.2in}
\end{table}

\section{Related Work} \label{sec:related-work}

\begin{figure*}[t!]
    \centering
\includegraphics[scale=0.49]{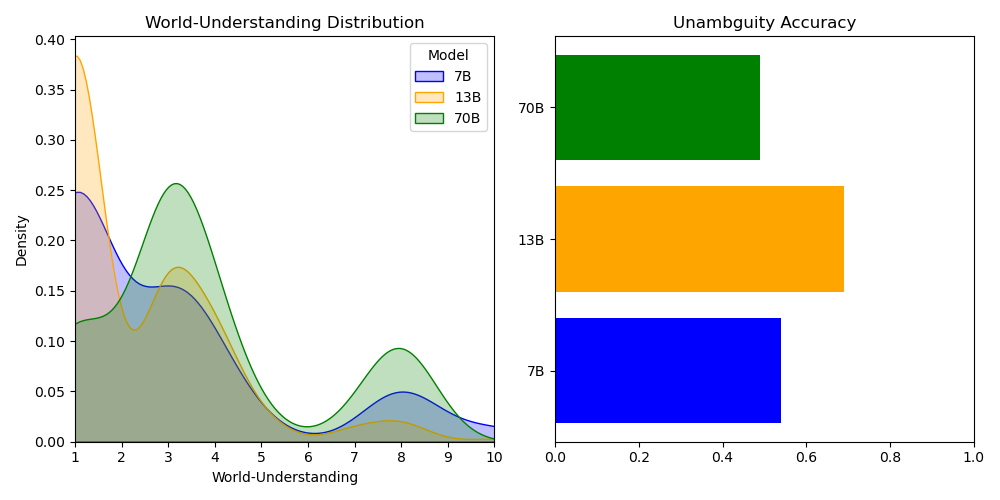}
    \caption{We report the qualities of the \glspl{iiu} generated using smaller, open-source Llama 2 models of three different sizes (7B, 13B, 70B). All the evaluation results are obtained using the best-performing GPT-4 proxy evaluation model (as described in Section \ref{sec:proxy-evaluators}).}
    \label{fig:llama-generation-results}
    \vspace*{-.2in}
\end{figure*}

\paragraph{Brittleness of \gls{dst} Models.} The initiative to develop the \gls{iiu} generation task springs from a need to reduce the brittleness of smaller \gls{nlu} and \gls{dst} models. \citet{cho_know_2022} empirically demonstrate the brittleness of commonly-used, small LM-based \gls{dst} models by showing that their performance degrades in the face of various types of perturbations involving linguistic variations, coreferences, named entity references, paraphrases, and speech disfluencies.
More generally, \citet{zarcone2021small} critique the academic community's prevailing focus on incremental advancements on synthetic benchmarks for tasks such as \gls{dst}, referred to as \textit{``playing the SNIPS game,''} which often overlooks deeper issues regarding dataset realism.

\paragraph{Relationship of \gls{iiu} Generation to Other NLP Tasks.}
\gls{iiu} generation is similar to paraphrase generation \citep{zhou2021paraphrase} in that both tasks are form of semantically-preserving text transformations. In fact, \gls{iiu} generation can be viewed as the task of generating a highly specific form of paraphrase (that adheres to our three linguistic criteria). It can also be viewed as the inverse of the \gls{nli} task, where the objective is to generate a premise entailing a given hypothesis, rather than inferring entailment from a premise-hypothesis pair, albeit in a different context from \citet{shen2018generating}. Most closely related to our work, \citet{ge2022should} propose linguistic criteria based on Gricean Maxims \citep{grice1975logic} for the task of generating follow-up questions for interactive surveys. While both tasks prioritize relevance and coherence, they differ in their objectives: the former aims to elicit information from the user, while the latter focuses on clarity and unambiguity in conveying requests, often serving as the initial turn or an independent subdialogue thread.

\paragraph{Text Generation using Small \glspl{llm}.} Our research also investigates the impact of model size on the quality of the generated \glspl{iiu}.
\citet{eldan_tinystories_2023} dispute the notion that smaller Language Models (LMs) inherently lack the capacity for intricate text generation tasks like storytelling. They attribute shortcomings to the 
prevalence of irrelevant information rather than model constraints. By assembling a targeted dataset of children's stories, they show that smaller LMs can produce narratives comparable to those by larger counterparts like GPT-3.5 and GPT-4. Our work is aligned with this broader spirit, aiming to match the output of a larger \gls{llm}s through fine-tuning a smaller model.



\section{Limitations and Future Work} \label{sec:limitations-future}

We have limited ourselves to supervised fine-tuning of \glspl{llm}. However, there is a rich literature 
on the use of reinforcement learning to guide language models towards specific text styles and content types, especially for abstract concepts of the likes of \textit{indirectness}, which can be explored as future work \cite{kaufmann2023survey}.

As \citet{bowman_what_2021} suggest, the ultimate evaluation measure for any NLP task should be grounded in in carefully annotated real user data. While modeling specific phenomena such as indirectness moves the needle on specific dialogue paradigms such as task-oriented dialogues, the community needs to evolve novel evaluation paradigms in the long run for wider forms of dialogue \citep{Mannekote2023TowardsAN}.

Finally, the linguistic criteria we have established for generating indirect requests in \iiutod\ are not only effective for the current dataset, but also serve as a robust and generalizable framework that can be leveraged in future work to create even more challenging and diverse datasets. For instance, by expanding the number of possible slot values per sample to tens or even hundreds, researchers can construct more complex and realistic datasets that push the boundaries of current \gls{nlu} models.

\section{Conclusion} \label{sec:conclusion}
In conclusion, our study addresses the gap between benchmark corpora and real-world utterances in task-oriented dialogue systems by focusing on the phenomenon of indirectness. We present a multi-stage \gls{llm}-based pipeline to generate \iiutod, a dataset of \glspl{iiu} based on the schemas from the \gls{sgd} dataset. \iiutod\ complements existing benchmarks, enabling the evaluation of \gls{nlu} and \gls{dst} models on realistic, indirect user requests that lack explicit slot values. Experiments with a state-of-the-art \gls{dst} model confirm the challenging nature of \iiutod. Furthermore, our data generation pipeline provides a versatile and efficient method for creating evaluation datasets for various task-oriented dialogue tasks on-the-fly, potentially driving significant improvements in the usability and performance of virtual assistants for the benefit of end users.


\bibliography{anthology_p2,custom,references,revisions}

\begin{thebibliography}{29}
\expandafter\ifx\csname natexlab\endcsname\relax\def\natexlab#1{#1}\fi

\bibitem[{Achiam et~al.(2023)Achiam, Adler, Agarwal, Ahmad, Akkaya, Aleman, Almeida, Altenschmidt, Altman, Anadkat et~al.}]{achiam2023gpt}
Josh Achiam, Steven Adler, Sandhini Agarwal, Lama Ahmad, Ilge Akkaya, Florencia~Leoni Aleman, Diogo Almeida, Janko Altenschmidt, Sam Altman, Shyamal Anadkat, et~al. 2023.
\newblock Gpt-4 technical report.
\newblock \emph{arXiv preprint arXiv:2303.08774}.

\bibitem[{Asri et~al.(2017)Asri, Schulz, Sharma, Zumer, Harris, Fine, Mehrotra, and Suleman}]{asri2017frames}
Layla~El Asri, Hannes Schulz, Shikhar Sharma, Jeremie Zumer, Justin Harris, Emery Fine, Rahul Mehrotra, and Kaheer Suleman. 2017.
\newblock Frames: a corpus for adding memory to goal-oriented dialogue systems.
\newblock \emph{arXiv preprint arXiv:1704.00057}.

\bibitem[{Blum-Kulka and Hamo(2011)}]{blum2011discourse}
Shoshana Blum-Kulka and Michal Hamo. 2011.
\newblock Discourse pragmatics.
\newblock \emph{Discourse studies: A multidisciplinary introduction}, 2(1):143--164.

\bibitem[{Bowman and Dahl(2021)}]{bowman_what_2021}
Samuel~R. Bowman and George Dahl. 2021.
\newblock \href {https://doi.org/10.18653/v1/2021.naacl-main.385} {What {Will} it {Take} to {Fix} {Benchmarking} in {Natural} {Language} {Understanding}?}
\newblock In \emph{Proceedings of the 2021 {Conference} of the {North} {American} {Chapter} of the {Association} for {Computational} {Linguistics}: {Human} {Language} {Technologies}}, pages 4843--4855, Online. Association for Computational Linguistics.

\bibitem[{Briggs and Scheutz(2017)}]{briggs2017strategies}
Gordon Briggs and Matthias Scheutz. 2017.
\newblock Strategies and mechanisms to enable dialogue agents to respond appropriately to indirect speech acts.
\newblock In \emph{2017 26th IEEE International Symposium on Robot and Human Interactive Communication (RO-MAN)}, pages 323--328. IEEE.

\bibitem[{Brown et~al.(2020)Brown, Mann, Ryder, Subbiah, Kaplan, Dhariwal, Neelakantan, Shyam, Sastry, Askell, and {others}}]{brown2020language}
Tom Brown, Benjamin Mann, Nick Ryder, Melanie Subbiah, Jared~D Kaplan, Prafulla Dhariwal, Arvind Neelakantan, Pranav Shyam, Girish Sastry, Amanda Askell, and {others}. 2020.
\newblock Language models are few-shot learners.
\newblock \emph{Advances in neural information processing systems}, 33:1877--1901.

\bibitem[{Budzianowski et~al.(2018)Budzianowski, Wen, Tseng, Casanueva, Ultes, Ramadan, and Gašić}]{budzianowski2018multiwoz}
Paweł Budzianowski, Tsung-Hsien Wen, Bo-Hsiang Tseng, Inigo Casanueva, Stefan Ultes, Osman Ramadan, and Milica Gašić. 2018.
\newblock {MultiWOZ}–a large-scale multi-domain wizard-of-oz dataset for task-oriented dialogue modelling.
\newblock \emph{arXiv preprint arXiv:1810.00278}.

\bibitem[{Cho et~al.(2022)Cho, Sankar, Lin, Sadagopan, Shayandeh, Celikyilmaz, May, and Beirami}]{cho_know_2022}
Hyundong Cho, Chinnadhurai Sankar, Christopher Lin, Kaushik~Ram Sadagopan, Shahin Shayandeh, Asli Celikyilmaz, Jonathan May, and Ahmad Beirami. 2022.
\newblock \href {https://doi.org/10.48550/arXiv.2112.08321} {Know {Thy} {Strengths}: {Comprehensive} {Dialogue} {State} {Tracking} {Diagnostics}}.
\newblock ArXiv:2112.08321 [cs].

\bibitem[{Eldan and Li(2023)}]{eldan_tinystories_2023}
Ronen Eldan and Yuanzhi Li. 2023.
\newblock \href {http://arxiv.org/abs/2305.07759} {{TinyStories}: {How} {Small} {Can} {Language} {Models} {Be} and {Still} {Speak} {Coherent} {English}?}
\newblock ArXiv:2305.07759 [cs].

\bibitem[{Ge et~al.(2022)Ge, Xiao, Diesner, Ji, Karahalios, and Sundaram}]{ge2022should}
Yubin Ge, Ziang Xiao, Jana Diesner, Heng Ji, Karrie Karahalios, and Hari Sundaram. 2022.
\newblock What should i ask: A knowledge-driven approach for follow-up questions generation in conversational surveys.
\newblock \emph{arXiv preprint arXiv:2205.10977}.

\bibitem[{Grice(1975)}]{grice1975logic}
Herbert~P Grice. 1975.
\newblock Logic and conversation.
\newblock In \emph{Speech acts}, pages 41--58. Brill.

\bibitem[{Hsieh et~al.(2023)Hsieh, Li, Yeh, Nakhost, Fujii, Ratner, Krishna, Lee, and Pfister}]{hsieh2023distilling}
Cheng-Yu Hsieh, Chun-Liang Li, Chih-Kuan Yeh, Hootan Nakhost, Yasuhisa Fujii, Alexander Ratner, Ranjay Krishna, Chen-Yu Lee, and Tomas Pfister. 2023.
\newblock Distilling step-by-step! outperforming larger language models with less training data and smaller model sizes.
\newblock \emph{arXiv preprint arXiv:2305.02301}.

\bibitem[{Karpinska et~al.(2021)Karpinska, Akoury, and Iyyer}]{karpinska-etal-2021-perils}
Marzena Karpinska, Nader Akoury, and Mohit Iyyer. 2021.
\newblock \href {https://doi.org/10.18653/v1/2021.emnlp-main.97} {The perils of using {M}echanical {T}urk to evaluate open-ended text generation}.
\newblock In \emph{Proceedings of the 2021 Conference on Empirical Methods in Natural Language Processing}, pages 1265--1285, Online and Punta Cana, Dominican Republic. Association for Computational Linguistics.

\bibitem[{Kaufmann et~al.(2023)Kaufmann, Weng, Bengs, and Hüllermeier}]{kaufmann2023survey}
Timo Kaufmann, Paul Weng, Viktor Bengs, and Eyke Hüllermeier. 2023.
\newblock A survey of reinforcement learning from human feedback.
\newblock ArXiv: 2312.14925 [cs.LG].

\bibitem[{Mann and Thompson(1988)}]{mann1988rhetorical}
William~C Mann and Sandra~A Thompson. 1988.
\newblock Rhetorical structure theory: Toward a functional theory of text organization.
\newblock \emph{Text-interdisciplinary Journal for the Study of Discourse}, 8(3):243--281.

\bibitem[{Mannekote(2023)}]{Mannekote2023TowardsAN}
Amogh Mannekote. 2023.
\newblock \href {https://api.semanticscholar.org/CorpusID:259950719} {Towards a neural era in dialogue management for collaboration: A literature survey}.
\newblock \emph{ArXiv}, abs/2307.09021.

\bibitem[{Mannekote et~al.(2023)Mannekote, Celepkolu, Wiggins, and Boyer}]{Mannekote2023ExploringUI}
Amogh Mannekote, Mehmet Celepkolu, Joseph~B. Wiggins, and Kristy~Elizabeth Boyer. 2023.
\newblock \href {https://api.semanticscholar.org/CorpusID:259938737} {Exploring usability issues in instruction-based and schema-based authoring of task-oriented dialogue agents}.
\newblock \emph{Proceedings of the 5th International Conference on Conversational User Interfaces}.

\bibitem[{Mavrina et~al.(2022)Mavrina, Szczuka, Strathmann, Bohnenkamp, Krämer, and Kopp}]{mavrina_alexa_2022}
Lina Mavrina, Jessica Szczuka, Clara Strathmann, Lisa~Michelle Bohnenkamp, Nicole Krämer, and Stefan Kopp. 2022.
\newblock \href {https://www.frontiersin.org/articles/10.3389/fcomp.2022.791704} {“{Alexa}, {You}'re {Really} {Stupid}”: {A} {Longitudinal} {Field} {Study} on {Communication} {Breakdowns} {Between} {Family} {Members} and a {Voice} {Assistant}}.
\newblock \emph{Frontiers in Computer Science}, 4.

\bibitem[{Mehri et~al.(2022)Mehri, Altun, and Eskenazi}]{mehri2022lad}
Shikib Mehri, Yasemin Altun, and Maxine Eskenazi. 2022.
\newblock Lad: Language models as data for zero-shot dialog.
\newblock \emph{arXiv preprint arXiv:2207.14393}.

\bibitem[{Rastogi et~al.(2020)Rastogi, Zang, Sunkara, Gupta, and Khaitan}]{rastogi2020towards}
Abhinav Rastogi, Xiaoxue Zang, Srinivas Sunkara, Raghav Gupta, and Pranav Khaitan. 2020.
\newblock Towards scalable multi-domain conversational agents: {The} schema-guided dialogue dataset.
\newblock In \emph{Proceedings of the {AAAI} conference on artificial intelligence}, volume~34, pages 8689--8696.
\newblock Issue: 05.

\bibitem[{Roberts et~al.(2019)Roberts, Raffel, Lee, Matena, Shazeer, Liu, Narang, Li, and Zhou}]{roberts2019exploring}
Adam Roberts, Colin Raffel, Katherine Lee, Michael Matena, Noam Shazeer, Peter~J Liu, Sharan Narang, Wei Li, and Yanqi Zhou. 2019.
\newblock Exploring the limits of transfer learning with a unified text-to-text transformer.
\newblock \emph{Google, Tech. Rep.}

\bibitem[{Samsi et~al.(2023)Samsi, Zhao, McDonald, Li, Michaleas, Jones, Bergeron, Kepner, Tiwari, and Gadepally}]{samsi2023words}
Siddharth Samsi, Dan Zhao, Joseph McDonald, Baolin Li, Adam Michaleas, Michael Jones, William Bergeron, Jeremy Kepner, Devesh Tiwari, and Vijay Gadepally. 2023.
\newblock From words to watts: Benchmarking the energy costs of large language model inference.
\newblock In \emph{2023 IEEE High Performance Extreme Computing Conference (HPEC)}, pages 1--9. IEEE.

\bibitem[{Sardana and Frankle(2023)}]{sardana2023beyond}
Nikhil Sardana and Jonathan Frankle. 2023.
\newblock Beyond chinchilla-optimal: Accounting for inference in language model scaling laws.
\newblock \emph{arXiv preprint arXiv:2401.00448}.

\bibitem[{Schegloff(1999)}]{schegloff1999discourse}
Emanuel~A Schegloff. 1999.
\newblock Discourse, pragmatics, conversation, analysis.
\newblock \emph{Discourse studies}, 1(4):405--435.

\bibitem[{Shen et~al.(2018)Shen, Tan, Huang, and Courville}]{shen2018generating}
Yikang Shen, Shawn Tan, Chin-Wei Huang, and Aaron Courville. 2018.
\newblock Generating contradictory, neutral, and entailing sentences.
\newblock \emph{arXiv preprint arXiv:1803.02710}.

\bibitem[{Touvron et~al.(2023)Touvron, Martin, Stone, Albert, Almahairi, Babaei, Bashlykov, Batra, Bhargava, Bhosale et~al.}]{touvron2023llama}
Hugo Touvron, Louis Martin, Kevin Stone, Peter Albert, Amjad Almahairi, Yasmine Babaei, Nikolay Bashlykov, Soumya Batra, Prajjwal Bhargava, Shruti Bhosale, et~al. 2023.
\newblock Llama 2: Open foundation and fine-tuned chat models.
\newblock \emph{arXiv preprint arXiv:2307.09288}.

\bibitem[{Wei et~al.(2022)Wei, Wang, Schuurmans, Bosma, Chi, Le, and Zhou}]{wei2022chain}
Jason Wei, Xuezhi Wang, Dale Schuurmans, Maarten Bosma, Ed~Chi, Quoc Le, and Denny Zhou. 2022.
\newblock Chain of thought prompting elicits reasoning in large language models.
\newblock \emph{arXiv preprint arXiv:2201.11903}.

\bibitem[{Zarcone et~al.(2021)Zarcone, Lehmann, and Habets}]{zarcone2021small}
Alessandra Zarcone, Jens Lehmann, and Emanu{\"e}l~AP Habets. 2021.
\newblock Small data in nlu: Proposals towards a data-centric approach.
\newblock In \emph{35th Conference on Neural Information Processing Systems (NeurIPS 2021)}.

\bibitem[{Zhou and Bhat(2021)}]{zhou2021paraphrase}
Jianing Zhou and Suma Bhat. 2021.
\newblock Paraphrase generation: A survey of the state of the art.
\newblock In \emph{Proceedings of the 2021 conference on empirical methods in natural language processing}, pages 5075--5086.

\end{thebibliography}
\bibliographystyle{acl_natbib}

\appendix

\section{Instructions shown to Human Annotators} \label{app:creativity-instructions}
For each task (sample), the annotators were required to fill in a form with two input fields. We provided examples along with brief instructions on how to fill in these fields (see Figure~\ref{fig:crowdsourcing-interface}) as shown below.

\textit{To get a feel for the task, please go through these examples.}
      
\textit{In all the examples below, the customer is trying to search for restaurants and indicating their preference for ``Italian cuisine.''}

\begin{enumerate}
  \item \textbf{Check all entailing slot values:} \textit{For the first question, you will need to check \textit{all} the values that can be implied by the customer's utterance. This could mean selecting zero, one, or more checkboxes. }[examples]
  \item \textbf{Use the slider to indicate the difficulty of the utterance.} [examples]
\end{enumerate}

\section{Prompts for Proxy Evaluators} \label{app:prompts-for-proxy-evaluator}

Below, we list the \gls{llm} prompts used for proxy evaluation of \unamb{} and \knowledge{} criteria.

\subsection{\unamb{}}
\begin{lstlisting}
    You are an expert at evaluating which slot value(s) could be implied by an utterance among a set of candidate values in a task-oriented dialogue. If no values can be eliminated, list all possible values separated by commas.
    Examples:
    Situation: User wants to make a trip
    Slot: Destination country
    Possible Values: India, Namibia, Nigeria
    Utterance: I'm looking to book a ticket to an African country
    Slot Values Implied: Namibia, Nigeria

    <more in-context examples>
\end{lstlisting}

\subsection{\knowledge{}}
\begin{lstlisting}
    On a scale of 1-10, how likely is it that an average six-year-old would be able to link the user utterance to the target slot value?
    Examples:
    Situation: User wants to find concerts and games happening in your area
    Slot: Destination country
    Possible Values: India, Namibia, Nigeria
    Utterance: I'm looking to book a ticket to an African country
    World Knowledge Level: 10

    <more in-context examples>
\end{lstlisting}







\section{Prompt for Generating \glspl{iiu}}
\label{app:iiu-generation-prompt}

Below is the prompt used to generate \glspl{iiu}.

\begin{lstlisting}
Generate a customer utterance containing an indirect and unique reason for wanting to choose a target slot value. Make sure that 1) the utterance entails ONLY the target slot value and that it DOES NOT mention the target slot value.

Situation: User wants to transfer money from one bank account to another user's account
Slot Description: The account type of the recipient whom the user is transfering money to
Possible Slot Values: checking, savings
Target Slot Value: checking
Do Not Mention: checking
Indirect User Request Keywords In: I need to transfer some money to my friend's account. He usually uses it for his direct deposits.

Situation: User wants to find a restaurant of a particular cuisine in a city
Slot Description: Price range for the restaurant
Possible Slot Values: inexpensive, moderate, expensive
Target Slot Value: moderate
Do Not Mention Keywords In: moderate
Indirect User Request: Looking to have a decent meal without burning a hole in my pocket

Now, generate ONE indirect user request for this input based on the above examples.
Situation: {situation}
Slot Description: {slot_description}
Possible Slot Values: {possible_slot_values}
Target Slot Value: {target_slot_value}
Do Not Mention Keywords In: {target_slot_value}
\end{lstlisting}





\section{Generation Parameters}
\paragraph{OpenAI Models.} We use the default settings from the OpenAI for our experiments with GPT-3.5 and GPT-4 models.

\paragraph{Llama 2 Models.} For all generation experiments with Llama 2, we use the following parameters.

\begin{description}
    \item[Top-k:] 50
    \item[Top-p:] 0.9
    \item[Temperature:] 0.5
    \item[Max New Tokens:] 128
    \item[Min New Tokens:] -1
    \item[Stop Sequences:] \texttt{\textbackslash n}
\end{description}

\end{document}